\documentclass[aps,prl,twocolumn,superscriptaddress,groupedaddress]{revtex4}
\usepackage{subfig}
\usepackage{graphicx}  
\usepackage{dcolumn}   
\usepackage{bm}        
\usepackage{amssymb}   
\usepackage{slashed}
\usepackage{graphicx}				
\usepackage{amsmath}
\usepackage{mathtools}
\usepackage{tikz,pgf}
\usepackage{comment}
\usepackage{asymptote}
\usetikzlibrary{arrows,backgrounds}
\usetikzlibrary{fit,scopes,calc,matrix,positioning,decorations.pathmorphing}
\usepackage[all]{xy}
\usepackage{yfonts}

\newcommand{\Bracket}[1]{\ensuremath{\left\langle#1\right\rangle}}

\begin{document}
\title{Are classical neural networks quantum?}
\author{Andrei T. Patrascu}
\address{ELI-NP, Horia Hulubei National Institute for R\&D in Physics and Nuclear Engineering, 30 Reactorului St, Bucharest-Magurele, 077125, Romania}
\begin{abstract}
Neural networks are being used to improve the probing of the state spaces of many particle systems as approximations to wavefunctions and in order to avoid the recurring sign problem of quantum monte-carlo. One may ask whether the usual classical neural networks have some actual hidden quantum properties that make them such suitable tools for a highly coupled quantum problem. I discuss here what makes a system quantum and to what extent we can interpret a neural network as having quantum remnants. 
\end{abstract}
\maketitle
Probably the best way of starting this note is to ask a question: what means to be quantum? 
Lacking a proper axiomatisation of quantum mechanics this question is particularly important because systems or mathematical structures we may associate to classical physics may have unexpected quantum properties. So, quantum mechanics has several defining principles, two of the most important being: its probabilistic nature: the outcomes generated by quantum mechanical calculations are always probabilistic, even if that probability gets very close to 0 or 1, and second, there is no pre-determined unique state of a system in the absence of an observable capable of asking a question about that property of the system. Essentially, that means the quantitative properties of a state of a system are not uniquely defined, and are not defined at all in the absence of an observable to be equated to a method of determining them. Moreover, not any observable is capable of fully determining the numerical value of a property of the state of a system, and hence some observables may not be compatible with respect to that property. This last property is somehow unfortunately called "non-realism" of the wavefunction, and that is unfortunate because there is nothing "unreal" about the wavefunction, in fact it provides the maximum of information about the system, and in many cases far more than a classical system could ever hope to achieve. As a consequence of this property, the rules of obtaining the probabilistic outcomes of quantum mechanics change in the sense that we have to employ a method of obtaining a module of a complex number, from a fundamentally complex quantity, the wavefunction, that needs to be complex in order to account for correlations in our knowledge about the system that will generate well known interference patterns in the probabilistic outcomes. 
That last property of quantum mechanics is Born's rule.
There exists a somehow widespread assumption that quantum phenomena must be either very small, or happen at low enough energy, or be small corrections of the order of $O(\hbar)$ to classical phenomena and that there is some form of "gap" between a so called "quantum world" and the "real" or "every-day" world. My thesis is that these assumptions are wrong. Of course, the approach to quantum mechanics phenomena initiated up to now are based on controlling the interference of the "catalog of knowledge" wave patterns (basically, a sort of wavefunction engineering), and those methods are extremely susceptible to quantum decoherence, namely extended interference of the environment which leads to an exponential increase of the terms arising in the "catalog of knowledge", to the point that the sheer complexity of the problem makes the returns impossible to identify in any meaningful sense. However, there are several macroscopic systems that are generally assumed to be classical, that by the way in which they work, have properties that give them some quantum properties, as far as the definitions above are concerned. One of these systems is the neural network. It is important to mention that there have been essays in which one speculated about the possibility of generating quantum computing processes in big enough biological neural networks by focusing on some detailed mechanism in some molecular substructure of the neural network [1]. I think indeed this approach is fundamentally wrong, or at least not feasible on a large scale (although one has discovered entanglement of spins in some animal brains [2,3]). Those methods are extremely susceptible to decoherence and are not expected to provide any more complex meaningful results. However, one may ask what if the large structure of a network has properties that at least in part reflect the fundamental axioms of quantum mechanics?
There have also been essays in constructing so called quantum neural networks [4]. Those networks have the links representing quantum processes implemented by unitary transformations, turning the neural network into a large quantum circuit. Those methods are promising and it would be amazing if they will be achieved. However, there is still the problem of decoherence that will require such networks to be extremely robust, unfortunately a property still beyond today's reach. It has however been noticed that classical neural networks do a great job at solving many body quantum problems with strong coupling [5], that are fundamentally quantum and expected to be properly solved on quantum computers. What are the properties of a classical neural network that would make it suitable for solving such complex quantum problems? 
This brief philosophical digression is here to underline what it means to be quantum, something that has indeed surprised early physicists but that should be quite obvious to us now. 
The mathematical underpinnings of such definitions are axiomatisation [6], that allows us to obtain the most abstract form of a concept, that can then be used to find it in various other forms and circumstances, usually not too obvious to a non-axiomatic approach, and categorification [7], that allows us to map categories containing certain groups and maps into other categories, apparently working in completely unrelated fields, but which act equivalently or in a way that can be related to the first category. If we try to do this with quantum mechanics we may notice that many structures, mathematical or computational, may have some level of quantumness, associated to some or all of the axioms, while being applied in completely different areas and seeming to a non-categorial mind as purely classical. 
A feature emerging from the quantum properties above is the inseparability of state spaces of quantum states [8]. This obstruction to cartesian pairing has been called entanglement, and it is advisable to keep that notion [9].
Entanglement is considered a characterising feature of quantum mechanics. It is a clearly non-classical phenomenon capable of offering correlations above any that could be obtained classically [10]. Entanglement is based on the non-separability of state spaces of systems that are connected to form an over-arching super-system. The non-separability relies on two concepts. First, taking two systems, each having their state spaces given, by combining them one obtains a paired state space with a dimension larger than the sum of the dimensions of the two subspaces, actually, the dimension will grow as the product of the dimensions of the two subspaces, leading to a pairing that involves significant global information that cannot be mapped or retrieved by local measurements on any of the two subsystems separately. The fact that pairing is not always cartesian is a feature of quantum mechanics that is linked to many quantum algebraic features. One can link this with the commutation relations that become non-abelian for incompatible observables, and finally with a probabilistic interpretation that allows, by means of statistics, to determine interference patterns linked to the global structure of the manifold on which the phenomena occur. In all cases, the two main aspects of the "quantumness" of a system are the linear product structure, and its ability to provide access to non-local features by means of a probabilistic, wave-function based interpretation. 
Much research is being performed in the direction of so called quantum neural networks. Those are networks in which the connecting edges are considered to be pieces of quantum circuits, leading to a network of quantum unitary operations being performed on a series of initial quantum states. There is no doubt that such a configuration would be highly useful from a practical point of view. The ability of performing incredible optimisation problems in this way would come into our reach. However, it is also well known that it is not at all easy to maintain coherence over a relatively large network of quantum states, which leads to (totally not insurmountable) practical problems. 
However, a classical neural network has certain properties that could be assimilated as quantum to some extent. While we would not propagate a quantum state through a series of quantum devices forming a hard to maintain network, we would use a set of classical neurons that, acting together, would be able to produce a state that could be classified as quantum for all practical purposes. Let us consider a neural network, made out of neurons firing according to classical rules. Let us also consider several layers of such neurons, as can be imagined in a deep learning network. The parameters of the neurons are bound together in a series of linear combinations in which the activation function plays the role of a non-linear contributor to the adaptability of the network. The relation among the neurons however is always linear. A signal is transmitted to the network as an input, which is passed then through the network and a gradient descent backpropagation method is used to train the linearly connected neurons forming the network to the desired structure. After defining a loss function, backpropagation is required to transmit the end-point error back to the input node. This process allows the network to access, via its non-linear activation function, the non-local structure of the features it explores. This is also why neural networks are so good at classifying non-local classification problems and are finally used as approximations for highly entangled quantum states. But are classical neural networks just approximations of quantum systems, or are they, in some sense, quantum themselves, not through the interactions that make them function, but through their global structure and their backpropagation features? The optimised result is indeed not easily associated with the input, which is why the artificial neural networks have the well known problem of inferring causal connections. 
In this sense, neural networks in general are devices capable of linking global and local information by using a method that reminds us of quantum correlations. Consider for example the situation describing reinforced learning. To do that we have to think of one or a set of agents interacting with an environment and developing a strategy resulting in the maximisation of a reward function. The maximisation problem is obviously non-local, and the agents are trained / learn global structures in their early training phase. The problem is stated usually as a Markov decision process (MDP) where we have a set of states $S$, a set of actions $A$, and the probability of a transition $P_{a}$ as well as a set of rewards given for executing an action $a_{t}$ given a state $s_{t}$. With no stochasticity, $P_{a}=1$ and we set up the goal for the agent to realise the policy $\pi(s_{t})=a_{t}$ such that we obtain a maximised reward 
\begin{equation}
E[\sum_{t=0}^{\infty}R(s_{t}, a_{t})|\pi]
\end{equation}
The policy is steadily learned, and we may introduce a discount parameter $\gamma$. The neural networks designed to learn the policy are constructed using parametrisations in the form weights and biases defined by $\theta$. The $Q$-value is given by 
\begin{equation}
Q(s_{t}, a_{t})=r_{t}+max_{a_{t+1}}Q(s_{t+1}, a_{t+1})
\end{equation}
and represents the numerical estimation of the reward after the agent performed the actio $a_{t}$ at the state $s_{t}$ and $r_{t}$ is the reward at the step $t$ while $max_{a_{t+1}}Q(s_{t}, a_{t})$ is the maximum future value of the reward. Our neural network is the $Q$-function and the learning policy is defined for discrete actions spaces being formulated as the policy parametrised by $\theta$ that would lead to the maximum $Q$ value. The Bellman equation is used to obtain the mean squared loss function and then to calculate the gradients needed for back-propagation
\begin{equation}
L_{t}(\theta)=E[(r_{t}+max_{a'}Q(s', a', \theta')-Q(s, a, \theta))^2]
\end{equation}
This is the standard DQN algorithm. Such an algorithm usually presents convergence problems because the maximisation can lead to over-estimations of $Q$. In order to correct that one may use separate target networks to predict the future $Q$ value inside the max operation, or dueling DQNs which have separate network heads that predict the advantage and value components of the Q-value, distributional DQNs, noisy nets, and others. 
In any case, the construction of the neural network, by means of a set of nodes linked by edges that provide inputs and outputs that are spread across a network have at least some properties which I want to underline here. 
First of all, the state produced by such a network is naturally non-separable. Hence it takes naturally into account one of the most fundamental aspects of entanglement, namely that the global system cannot be separated into its components without taking into account some form of common shared global information. 
Let us consider a neural network defined by an activation function $\sigma : \mathbb{R}\rightarrow \mathbb{R}$. This activation function may be a vector $\sigma(x)=(\sigma(x_{1}), \sigma(x_{2}), ... , \sigma(x_{3}))$. Let us consider a function $f:\mathbb{S}^{d-1}\times \mathbb{S}^{d-1}\rightarrow \mathbb{R}$ such that $f(x,x')=g(\Bracket{x,x'})$, for $g:[-1,1]\rightarrow R$. It has been shown [2] that the depth of the neural network gives a profound difference in whether the function can be approximated by a neural network. Indeed, $F:\mathbb{S}^{d-1}\times \mathbb{S}^{d-1}\rightarrow \mathbb{R}$ can be implemented by a depth-$2$ network of width $r$ and weights bounded by $B$ if 
\begin{equation}
F(x,x')=w_{2}^{T}\sigma(W_{1}x+W_{1}'x'+b_{1})+b_{2}
\end{equation}
with $W_{1},W_{1}'\in [-B,B]^{r\times d}$, $w_{2}\in[-B,B]^{r}$, $b_{1}\in[-B,B]^{r}$ and $b_{2}\in[-B,B]$. 
$F$ could be implemented by a depth-$3$ $\sigma$-network of width $r$ and weights bounded by $B$ if 
\begin{equation}
F(x, x')=w_{3}^{T}\sigma(w_{2}\sigma(W_{1}x+W_{1}'x'+b_{1})+b_{2})+b_{3}
\end{equation}
for $W_{1}$, $W_{1}'$ $\in [-B,B]^{r\times d}$, $W_{2}\in [-B,B]^{r\times r}$, $w_{3}\in[-B, B]^{r}$, $b_{1}, b_{2} \in [-B, B]^{r}$ and $b_{3}\in[-B,B]$. 
Polynomial-size depth two neural networks with exponentially bounded weights will not be able to approximate $f$ whenever $g$ cannot be approximated by a low degree polynomial, however such functions can be approximated by polynomial size depth three networks with polynomial bounded weights. This gives a fundamental non-linear distinction between a depth $2$ and a depth $3$ neural network. This result however is not sufficient to prove non-separability. In order to do that, one has to also consider the agent-based learning process and the maximisation principle. This optimisation depends non-linearly on the possible cuts one could make in the network. If one reduces depth of a network one cuts through a tensorial product of the network components, leaving us with the results properly encoded by the distinction between a cartesian and a tensorial product. 
In fact it has been shown that a neural network with a single hidden layer with sigmoid activation functions and additional non-linearity in the output Neuron can learn ball indicator functions efficiently while using a reduction technique ball indicators cannot be approximated efficiently using depth-2 neural networks when the non-linearity in the output neuron is removed [3]. Therefore in some cases a larger neural network lacking some non-linearity in the output becomes less capable at learning functions than a smaller network with some output non-linearity. 
The depth of the network can be described as follows. Let us have a network in the form of a sequence of layers $\{L_{0}, L_{1}, ..., L_{N}\}$. The input is being provided at one side of the layers, and is transferred through the network which computes it as a sequential application of the layers
\begin{equation}
F(data)\leftarrow L_{N}(L_{N-1}...(L_{0}(data)))
\end{equation}
The loss function will be used to compute the gradients for the final layer $G_{loss}(output, label)$. The backpropagation over one layer will therefore be $L_{i}(\nabla)$ while the backpropagation over the entire network will be denoted $F^{T}(\nabla)$. Backpropagation over the entire network is represented by a sequential application of backward layers 
\begin{equation}
F^{T}(\nabla) \leftarrow L_{1}(L_{2}(T)... (L_{N}^{T}(\nabla)))
\end{equation}
The process of backpropagation allows the network to construct the suitable internal representations given by its sets of parameters, such that it can learn the mapping connecting the input to the output. The main part of the backpropagation method is the construction of the gradient of the loss function with respect to the weights of the network. The output of each neuron goes through an activation function, say $\sigma(x)=\frac{1}{1+e^{-x}}$ and combines all the previous results of the neural net, namely $\sum_{k=1}^{n}w_{kj}o_{k}$. As the derivative of the activation function is $\sigma'(x)=\sigma(x)(1-\sigma(x))$ the input of a neuron usually includes the weighted output of all the other previous nerons. 
If we think about this process, we will find some rather interesting connections to what we learn from quantum mechanics. Particularly, the way in which global information is being dealt with reminds us of quantum superposition of various forms. The only difference is the way we choose to interpret those outcomes. In fact, often the outcomes of a classifying neural network are probabilistic, in the sense that they provide probabilities for the correctness of the outcome of the classification problem. However, if we look at the internal mechanisms of the neural network, we notice a series of linear combinations followed by integration through the input branches of neurons that lead to superposed effects in the output. Each neuron comes with a non-linear component, namely the activation function, which allows the neural network to learn and take into account non-linear effects. What we obtain is a global characterisation of the state of each neuron in the form of a linear superposition, modulated by a non-linear activation function which introduces a threshold for the superposition to be considered. 
The backpropagation sends the information about the optimisation results back to the network, tracing back the layers and allowing for updates. This step is obviously crucial as, without it, no learning could take place. This inverse problem is actually one of the most investigated and optimised tool in machine learning. It is exceptionally important for the gradient descent methods to be well designed to make this process amenable to a rather large problem as the one generated by neural networks. However, another aspect of this method is to reinforce the non-separability of the network. Depending on the network depth, we have different learning capacities. The whole process of forward and back - propagation amounts to what we would call in quantum mechanics entanglement. Indeed, from this point of view, the classical neural network is a highly entangled state. How can that be even possible if the entanglement is basically a quantum property that relies on the non-realism principle described above? There could be no entanglement in systems showing only one single underlying fundamental outcome. However, we already went through the process of generating the neural network, and it appeared in several situations how this assumption stops being true. For any given input state, which finally is classical, as in, determined with one absolute value for its properties, the neural network does something interesting. It takes those input values, it generates a set of linear superposition between them, and it applies them to the next layer by means of activation functions. The activation functions cut some signals, while allowing others to pass, and the optimisation algorithm forces the network to gain access to the global structure of the problem manifold. Therefore, in the training phase, we are faced with a series of potential states of the system, each taken into account in each iteration of the loss function, and each being forced to take into account the global structure of our problem. This provides our network with a series of many potential outcomes, hence it lifts its state from one single value for its properties, to several possible values. Therefore, we could, in principle, describe the network as a whole by means of an operator-valued observable. We can also generate its potential eigenstates, by following closely the learning process. This still does not make the system truly quantum. What we need is non-separability. This is provided by the back-propagation of the gradient to make the optimisation/learning possible. At each step when the network learns, and therefore it submits its end-point parameters backwards through the network, it generates global information by means of its loss function, that could not be found at the previous step, when the signal did not yet advance up to the current layer. 
Of course there is the final step, which in quantum mechanics is the Born rule. In quantum mechanics we deal with complex wavefunctions representing expectation catalogues (expression I learned from Schrodinger, and I enjoy using it to describe what wavefunctions are). Those wavefunctions have complex phases that allow them to combine and form wave-patterns leading to the famous problems of quantum interference in double slits experiments, etc. To obtain the probability (density) out of those, we need to apply Born's rule, namely to calculate the absolute value or the norm of the final combination of complex wavefunctions. Neural networks also combine various branches of their inputs, leading to interesting interferences, but they do that by branching out "dendrites" for input signals and providing outputs. The outputs will really depend on how we decide to interpret those combinations. In most applications to strongly entangled quantum problems, those inner branches of a neural network are being interpreted as individual wavefunctions, and hence the process of combination inherent to a neural network, will construct proper entanglement which allows us to obtain accurate simulations of our quantum system. If however, we force the neural network to operate with classical probabilities, we will obtain classical results. Finally, the sole difference between a neural network and a quantum system is the way we decide to deal with the representation of data that we feed to the network and with the output. If what we feed to the network is a wavefunction, and we treat it as such, the neural network will optimise it in a quantum manner, giving a decisively better approximation to the quantum problem at hand. Of course, the process should not be mistaken for quantum computation. A quantum neural network proper, would do a far better job as it would harness the real power of entanglement for each of the branches, which would become quantum circuit lines. However, while not being a quantum computer, the neural network certainly has some remnant quantum properties that are inherent to its working style. 
Probably the most important aspect of this way of thinking is to understand what it means to be quantum. The non-separability of a space of states is essential, together with the linearity of the processes. The linear combinations are being performed by the neural network and integrated, but the process of interference is done via a non-linear function, the activation function. However, aside of that, the process has many similarities with a quantum system. Particularly, it generates entanglement, in the sense of generating non-separable state spaces on the network. this is being done by means of the back-propagation and learning sections of the method. This allows us to access global information of the problem manifold. That information would not be retrievable in any separated piece of the network. This also explains some of the standing problems of neural networks, namely that it is usually hard to impossible to detect causal relations between the inputs and the outputs of the network, and that it is generally difficult to formulate local explanations of the methods leading to the outcome of the learning phase. This is so precisely because there is no local way of understanding the global problem, nor is there a strict causal connection, rather one that is based on weak-quantum correlations, at least not in most cases, exceptions are known. 
\par Let us look at the process described above by means of a simple system with one input neuron, one hidden neuron, and one output neuron. In the first phase we have the transition from our initial neuron to the hidden neuron, the information is being combined via a linear combination, and the non-linear activation function generates the output, and gives us access to its own parameters. The hidden neuron then carries the result of the activation function as an output towards the input of the next neuron or layer. There another linear combination is performed, another non-linear activation function is introduced and the output is generated. If there was only this step of forward generation and propagation, then the neural network would be separable (at least in principle). Sure, reducing its size would affect the outcomes, but the global information would emerge only at the level of a one-step linear combination. However, we also have the learning phase. That particular phase generates the back-reaction where the gradient is being propagated in the direction of the blue wave-fronts. The parameters are being transferred by means of this gradient backwards via the blue arrows and a learning process is started. This phase introduces global information through the loss function that makes the neural network not separable, and makes the global state manifest, a state that cannot be recovered in the local neurons anymore. In quantum mechanics we call this entanglement. 
\begin{figure}
  \includegraphics[width=\linewidth]{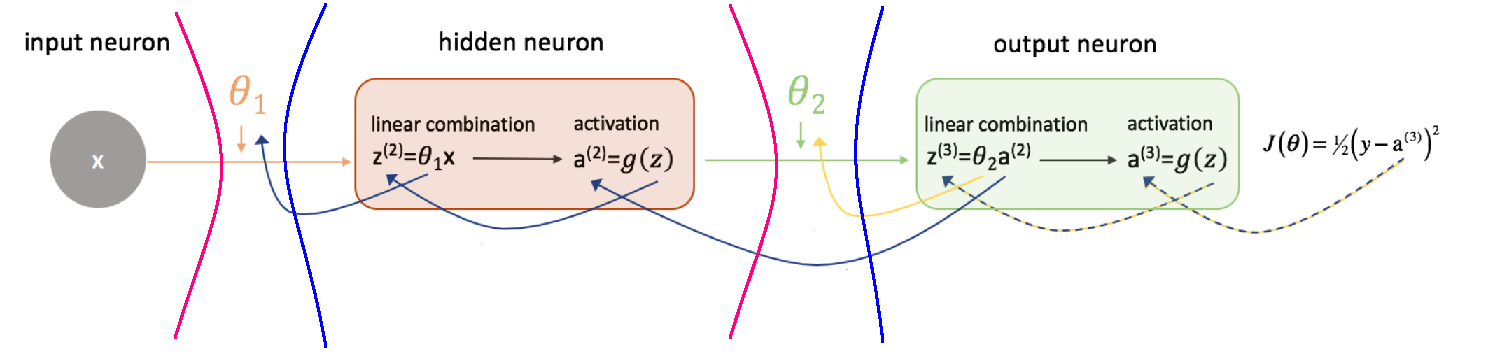}
  \caption{The evolution of information in a simple neural network. The red wavefront shows the direction in which information is separable. The gradient back-propagation however, drawn as the blue wavefront makes the global information on the network non-separable and the overall state space of the network more "quantum"}
  \label{fig:gauge1}
\end{figure}
The way in which this neural network is quantum is probably not the most obvious one, or one that is expected by physicists working with quantum mechanics at a different level. If one axiomatises quantum mechanics and generates a series of axioms that could define it completely as such, one may ask what would happen if some of those axioms are being abandoned? Quantum mechanics has its very special way of dealing with global information, namely it postulates, correctly so, that there exists global information that cannot be recovered locally. This is the foundation of entanglement. It also postulates the existence of catalogues of maximal knowledge (our wavefunctions) that can pre-interfere, before their results are being determined by properly constructed non-ambiguous observables. Both these aspects are being recovered in classical neural networks. The last one aspect that is not inherently built in the neural networks is the interpretation of the input and output information. In quantum mechanics we know the input information needs to be in the form of a complex expectation catalogue, while to obtain relevant output, we need to perform a Born-type procedure that will result in the correct probabilistic answer. It appears like quantum mechanics without this last axiom is found in neural networks. 

\end{document}